# Frequency Domain Convolutional Neural Network: Accelerated CNN for Large Diabetic Retinopathy Image Classification

Ee Fey Goh[a], ZhiYuan Chen[a,*], Wei Xiang Lim[a]

[a] University of Nottingham Malaysia, School of Computer Science, Jln Broga, 43500 Semenyih, Selangor

**Abstract**

The conventional spatial convolution layers in the Convolutional Neural Networks (CNNs) are computationally expensive at the point where the training time could take days unless the number of layers, the number of training images or the size of the training images are reduced. The image size of 256x256 pixels is commonly used for most of the applications of CNN, but this image size is too small for applications like Diabetic Retinopathy (DR) classification where the image details are important for accurate classification. This research proposed Frequency Domain Convolution (FDC) and Frequency Domain Pooling (FDP) layers which were built with RFFT, kernel initialization strategy, convolution artifact removal and Channel Independent Convolution (CIC) to replace the conventional convolution and pooling layers. The FDC and FDP layers are used to build a Frequency Domain Convolutional Neural Network (FDCNN) to accelerate the training of large images for DR classification. The Full FDC layer is an extension of the FDC layer to allow direct use in conventional CNNs, it is also used to modify the VGG16 architecture. FDCNN is shown to be at least 54.21% faster and 70.74% more memory efficient compared to an equivalent CNN architecture. The modified VGG16 architecture with Full FDC layer is reported to achieve a shorter training time and a higher accuracy at 95.63% compared to the original VGG16 architecture for DR classification.

## 1 Introduction

Diabetic Retinopathy (DR) is a chronic eye disease commonly found in diabetic patients and it is the most frequent cause of blindness. It is estimated that 0.8 million people were blind and 3.7 million were visually impaired due to DR, the number was increased by 27% and 64% respectively from 1990 to 2010 [1]. The high level of blood sugar can damage the retinal capillaries, causing capillary leakage and blockage which can lead to blindness [2]. DR can be detected by the appearance of features including microaneurysms, haemorrhages, soft exudates and hard exudates. In the early stages of DR, the high level of sugar in blood causes dilation in the blood vessels, which can result in the formation of blood-filled bulges called microaneurysms [2]. Microaneurysms are early signs of DR, it appears as small red dots on the fundus image. The microaneurysms may also bleed or leak into the retina, which is then called hemorrhages. As the disease progresses, many blood vessels will be blocked and lose the ability to transfer blood, the blockage then can form cotton wool spots, also called soft exudates. The breakdown of the retina vessels then causes leakage of proteins and lipids which forms hard exudate. Therefore, the early detection of DR is vital, and there has been a continuous effort to make DR detection more accurate and feasible through medical image analysis.

The use of Convolutional Neural Networks (CNNs) for image classification has been successful in real-world applications including DR detection [3]. However, the training of a CNN can take days depending on the size and amount of the data images [4], due to the computationally expensive convolution operations. As a result, the input images are commonly downscaled to 256x256 pixels or even smaller, which may be suitable for tasks like classifying between cats and dogs, but not for DR diagnosis [5]. The reason is that the details in retinal images are important for DR diagnosis, that is why the retinal images are usually taken from high resolution cameras. It would be a great loss of information if it was downscaled to 256x256 pixels because capillary dilatations like microaneurysms are tiny and may be lost after such significant image downscaling.

* Corresponding Author: ZhiYuan Chen (Zhiyuan.Chen@nottingham.edu.my)

In order to speed up CNNs to be able to train on larger images efficiently, researchers including Mathieu et al. [6] and Rippel et al. [7] tried to construct frequency domain convolution based on the convolution theorem, which shown to reduce the time complexity from $O(N^2)$ to $O(NlogN)$ for each convolution. Early results of frequency domain convolution have shown potential for further improvements, however, the research on frequency domain CNN and frequency domain components are severely lacking. Therefore, the focus of the research is to further explore and improve the existing frequency domain components including frequency domain convolution and frequency domain pooling. The frequency domain components are then used to improve the existing CNN architectures and to build a frequency domain CNN architecture for DR classification.

## 2 Materials and Methods
### 2.1 *Area of the research work*
The dataset for the training and testing of all the CNN architectures in this research is the APTOS 2019 blindness detection competition [8] dataset hosted on Kaggle. The dataset contains a total of 3662 high quality fundus images labelled with 5 severity classes of DR. The dataset contains images with varying size and color temperature, but overall, the images are clear and of good quality. The difference in color temperature of the fundus images may be due to different lightning equipment or fundus camera, the experiments are decided to work with grayscale fundus images to eliminate the color temperature difference and to focus on the details of the fundus. Additionally, due to the heavy class imbalance, the research will be focused on binary class classification by combining the "Mild", "Moderate", "Severe" and "Proliferative" class as the positive class, to be compared with the "No DR" negative class.

On a side note, there is a complication during the implementation of FDCNN (defined in Section 2.5.1) that makes it rely on the correct batch number to function properly. A temporary solution is done by trimming the dataset so that it will be divisible by the batch size. For the experiments in this dissertation, a batch size of 4 is used. Thus, 2 images from class 2 are randomly removed so that the dataset is divisible by 4 after splitting. The dataset is split into a 60% training set, 20% validation set and 20% testing set, summarized as in Table 1:

**Table 1** Dataset distribution for the experiments.

| Class    | Training images | Validation images | Testing images |
|----------|-----------------|-------------------|----------------|
| 0 - No DR | 1083            | 361               | 361            |
| 1 - DR    | 1113            | 371               | 371            |

### 2.2 *Image Preprocessing and Augmentation*
The images in the APTOS 2019 dataset are originally colored rectangle images with the background in black. First, the images are converted into grayscale, then the thresholding method is used to create a mask that separates the fundus area into white, while the background stays as black. The images are crop tight according to the white area of the mask, then downscaled to 512x512 pixels. After that, it performs Contrast Limited Adaptive Histogram Equalization (CLAHE) to boost the contrast of the image. The image augmentation step performs random rotations, horizontal and vertical flip on the images for every epoch to artificially create variations of the training image.

### 2.3 *Convolutional Neural Networks (CNNs)*
The Convolutional Neural Networks (CNNs) have several components such as the convolution, pooling, activation function and fully connected layers. The convolution operation in CNNs works in the same fashion as in image filtering in image processing, which is conducted by addition on the surrounding pixels (defined by kernel size), with each pixel multiplied to the corresponding kernel weights, defined as

$$O(i,j) = \sum_{k=1}^{n}\sum_{l=1}^{n} I(i+k-1, j+l-1) \cdot K(k,l). \quad (1)$$

Assume a square image of width $N$ and a square kernel of width $n$, the convoluted output image $O$ will be smaller with $i$ and $j$ both ranging from 1 to $N - n + 1$. Different padding techniques can be used on the edge of the input image to retain the image size on the output image, but it may also introduce artifacts. Convolution is the most computationally expensive operation in CNNs, it has a time complexity of $O(N^2)$ when it is performed on a single channel image. The convolution layers used in CNNs are usually performed over multiple channels of the image with different sets of kernels, described as Convolutions Over Volume (COV). First, each image channel is convoluted with each kernel channel, then all convoluted channels are summed together into a single channel image, the process is repeated over different sets of kernels to produce a multiple channel image, described as

$$COV = \bigcup_{s=1}^{S} \sum_{c=1}^{C} I_c * K_{sc} \quad (2)$$

where $*$ denotes convolution described in Equation 1, $c$ denotes the channel index, $C$ denotes the number of image channels, $\cup$ denotes concatenation, $s$ denotes the set index and $S$ denotes the number of sets or the number of channels to produce in the final image. The total number of weights in a convolution layer which uses COV is described as

$$K_{weights} = S \cdot C \cdot K_{height} \cdot K_{width}. \quad (3)$$

### 2.3.1 *Application of CNNs for DR classification*

Xu et al. [4] proposed an 8 convolution layer CNN architecture for binary class DR classification. Their model was trained on only 800 color fundus images. Although they achieved an accuracy of 94.5%, it was only tested on a 200 color fundus images. They downscaled the fundus images to 224x224 pixels for training and reported that the training process took 2 days with a NVIDIA GTX 1070 GPU. Although they were using the Kaggle 2015 DR dataset which has 80,000 images, they were not able utilize much of the data from the dataset due to the time consuming process of training the CNN.

Lam et al. [5] utilized transfer learning on a pretrained GoogLeNet CNN architecture. It was found that the use of Contrast Limited Adaptive Histogram Equalization (CLAHE) for image preprocessing and the use of Adam optimizer achieved the best accuracy of 74.5% for binary classification. Their model was trained on 1000 images and tested on 400 images, both at the size of 256x256 pixels. It was trained over 30 epochs, and each epoch required about 20 minutes on a NVIDIA TESLA K80 GPU. Lam et al. [5] also reported an accuracy of only 57.3% by applying GoogLeNet for multi-class DR classification, they suggested that the model had a difficult time differentiating apart multiple classes due to the extreme image downscaling, which the important fundus features were lost.

Model ensemble has been widely used in recent years. Wang et al. [9] proposed to ensemble CNN and random forest classifiers for retinal blood vessel segmentation. In the Kaggle 2019 DR competition, the top winners have all reported to use ensemble on at least 6 CNN models. Pratt [10] made use of a similar concept of ensemble and proposed the PatientDenseNet architecture. PatientDenseNet was made of 2 parallel DenseNet121, each for extracting the features of the left and right eye, then their outputs were connected to the fully connected layers. The experiment was run on the full Kaggle 2015 DR dataset and achieved an accuracy of 88% over 80 epochs which took 1240 minutes.

### 2.3.2 *Development in Frequency Domain Based CNNs*

Mathieu et al. [6] were the first to explore the use of Fourier transform and convolution theorem in CNNs. In their implementation of frequency domain convolution, there is a need for constant switching between the spatial and frequency domain with Fourier transforms before and after each convolution, which became an expensive computation overhead. Throughout the years, the frequency domain based CNNs have been ignored and underexplored due to the lack of frequency domain components especially the nonlinear activation function. Vasilyev [11] demonstrated that the ReLU activation function would become convolution operation in the frequency domain, making the computational complexity high and thus losing the point of using frequency domain CNNs. Similarly, Ayat et al. [12] makes use of the principle but approximated ReLU by a parabola and proposed SReLU. Recently, Watanabe and Wolf [13] proposed the Second Harmonics SReLU (2SReLU) which was an attempt in efficient frequency domain activation function by adding the lower frequencies back into the image to approximate the effect of ReLU. For the pooling layer, Rippel et al. [7] proposed the Spectral Pooling method for frequency domain CNNs which was based on the low pass filter. They showed Spectral Pooling has an advantage of being able to retain more spatial information compared to conventional pooling. Furthermore, Wang et al. [14] combined frequency domain convolution and Spectral Pooling into a single layer and showed that it can be efficient as input data images increase in size.

Pratt [10] claimed to be the first to propose a CNN that fully stays in the frequency domain which he called Fourier CNN (FCNN). The FCNN also made use of frequency domain convolution and Spectral Pooling, while Fourier transform was only required to be performed once in the beginning. However, there were several important details not being described in Pratt [10], especially that the implementation details of a nonlinear activation function in the frequency domain was missing, despite the fact that it has been the biggest obstacle in achieving a fully frequency domain CNN.

### 2.4 *Frequency Domain Components*
The term frequency domain refers to the representation of images as frequencies, instead of pixels which are used in the conventional spatial domain [15]. Fourier transforms are performed in order to transform images into the frequency domain and back into the spatial domain, which can be computationally expensive in repeated use. Therefore, there is a need for the equivalent frequency domain components to reduce the use of Fourier transforms.

#### 2.4.1 *Frequency Domain Convolution (FDC)*
The term frequency domain convolution is used to refer to the pointwise product of frequency domain images based on Convolution Theorem. While FDC is used to refer to the whole proposed implementation.

The core of frequency domain convolution is made possible by the Convolution Theorem. Let $I$ and $K$ be a single channel image and a single channel kernel, $F$ as the Fourier Transform function, the Convolution Theorem states that the Fourier transform on the spatial convolution (denoted as $*$, defined in Equation 1) is equal to the pointwise product (denoted as $\circ$, also called as Hadamard product) in the frequency domain, defined as

$$F(I * K) = F(I) \circ F(K). \qquad (4)$$

The advantage of pointwise product is that the time complexity is $O(N)$. However, the time complexity of Fourier transform is $O(NlogN)$ with the commonly used Fast Fourier Transform (FFT) algorithm. Therefore, convolutions in the frequency domain have a time complexity of $O(NlogN)$, which is still more efficient than spatial convolution with $O(N^2)$.

#### 2.4.1.1 *Real-to-Complex Fast Fourier Transform (RFFT)*
Fast Fourier Transform (FFT) is a general algorithm that computes the Discrete Fourier Transform (DFT) of an input sequence, which converts the input signal from time or spatial domain into the frequency domain. By taking advantage of the fact that images in the spatial domain are always real numbers, the images when represented in the frequency domain will be a Hermitian symmetric with the negative frequencies equal to the complex conjugates of the corresponding positive frequencies [16]. In result, RFFT reduced the width of the image to $N/2 + 1$ while the height maintains at $N$. The variant of FFT that makes use of the Hermitian symmetric property is known as RFFT.

### 2.4.1.2 *Kernel Initialization Strategy*
In order to minimize the transformation of kernels between both domains, Pratt [10] proposed to initialize all the kernels directly in the frequency domain using Xavier initialization. However, the initialization strategy inevitably lost the ability to have control over the kernel size as the kernels have to be directly the same size as the image. As the size of kernels affects the strength of the convolution, the kernels in Pratt [10] would be overpowered, not to mention it would also significantly increase the number of weights in the convolution layer.

The proposed FDC layer will have the kernels initialized in the spatial domain for the ability to define kernel sizes. Every time when the FDC layer is used, the kernels are transformed into the frequency domain to perform convolution with the frequency domain image. The kernels are not needed to be transformed back into the spatial domain after convolution. The advantage is that the kernels can be treated the same as in conventional convolution layers, so that it can be initialized with Kaiming [17] initialization, and the backpropagation algorithm is not needed to be modified to work with complex numbers.

### 2.4.1.3 *Convolution Artifact*
The convoluted image with spatial convolution will result in a smaller image depending on the size of the kernel. The only way to retain the size of the image after spatial convolution is to pad the image with $\frac{n-1}{2}$ pixels on all sides, some padding techniques like duplicating the edge values can work in some types of images, but generally the padded image will result in artifacts on the edges. Similarly, convolutions in the frequency domain possess edge artifacts, but paddings are not needed as frequency domain convolutions are pointwise products. The difference between spatial and frequency domain convolution is that the artifacts on frequency domain convolution are on the top and left side of the image, but it can be shifted with $\frac{n-1}{2}$ pixels to distribute to all edges.

The decision on whether to pad the image on spatial convolution is a design choice that varies with the CNN architecture. For frequency domain convolution, in the case where the image does not need to maintain the same size after convolution, the artifacts can be removed by cutting $n - 1$ pixels on the top and left side of the image, the resulting size of the image will be the same as spatial convolution without paddings.

Compared to the FCNN architecture from Pratt [10], as the images are never converted back into the spatial domain, the convolution artifacts can't be removed, which can possibly harm the learning of the CNN. The convolution artifacts problem was also not addressed in any previous works on frequency domain based CNNs.

### 2.4.1.4 *Channel Independent Convolution (CIC)*
Until now frequency domain convolutions are described only on a single channel image and kernel. As described in Section 2.3, the spatial convolution layer performs convolutions with different sets of kernels over all the channels, also referred as Convolutions Over Volume (COV). The COV method is computationally expensive when it comes to a large number of channels as all the convoluted channels within a set of kernels are summed into a single channel image. With the aim of improving the speed of the FDC layer for feature extraction through convolutions, the method Channel Independent Convolution (CIC) is proposed to minimize the use of convolutions while being able to produce the same number of channels compared to COV. The CIC method treats each channel like an independent image and the convolutions performed on each channel are not summed together, so it requires a smaller number of kernels to produce a convoluted image. The CIC method is defined as

$$CIC_c = \cup_{j=1}^{J} I_c \circ K_{cj} \qquad (5)$$

where $c$ denotes the input channel index which ranges from 1 to all the input channels, $\circ$ denotes pointwise product or frequency domain convolution, $\cup$ denotes concatenation and $J$ is the number of single channel kernels to be convolved with each single channel image. An illustration can be seen on Fig. 1 where $J = 2$.

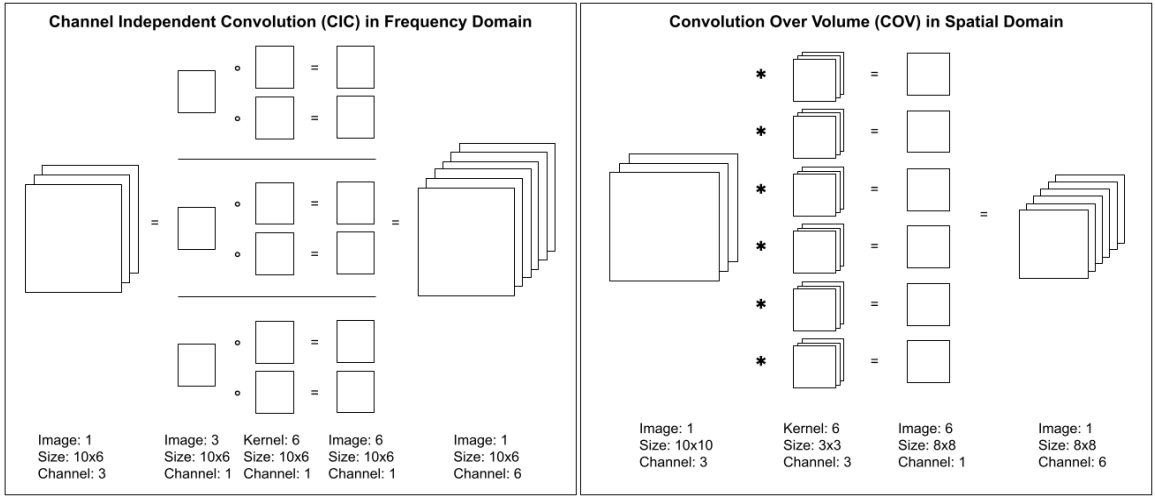

**Fig. 1** Comparison between CIC and COV method. Note that the image size after CIC will be the same as COV after transformed back into the spatial domain with convolution artifacts removed.

As CIC performs less convolutions than COV, the number of kernels required is also reduced. Additionally, the CIC method will also result in a smaller number of weights. As the images in CIC are split into single channel images, Equation 3 can be adjusted with $C = 1$ to describe the weights in FDC layer which uses CIC, shown as

$$K_{weights} = S \cdot K_{height} \cdot K_{width} \qquad (6)$$

where $S$ is the number of sets of kernels or the desired number of channels to be produced. However, unlike COV where $S$ can be any integer, the $S$ value in CIC has to fulfill $S \bmod C = 0$ where $C$ is the number of input channels. With that in mind, the $J$ factor from Equation 5 can also be calculated by $J = \frac{S}{C}$. The number of weights in CIC is reduced by a factor of $C$, while CIC also performs $S(C - 1)$ less convolutions than COV. However, as indicated previously, the downside of CIC is the $S$ value constraint. Fortunately, the $S$ value constraint is usually fulfilled even with spatial convolution layers, as it is common to set the output channels ($S$ value) to be exactly doubled of the input channels ($C$ value).

#### 2.4.1.5 *FDC Layer Implementation*
The FDC layer combines the details that were discussed previously and its implementation is shown as below:

**Algorithm 1**: FDC layer

**Input:** Images with $C$ channels in frequency domain,
    $S$ number of $K_{height} \cdot K_{width}$ kernels in spatial domain
**Output:** Convoluted $C \cdot J$ channel images in frequency domain
**procedure** FDC layer

    Pad kernels to the same size as images
    Convert kernels into frequency domain with RFFT
    Calculate $J$ factor with $J = \frac{S}{C}$
    Perform Channel Independent Convolution
    **return** convoluted $C \cdot J$ channel images
**end procedure**

        The FDC layer is further wrapped in a class that extends the PyTorch class to enable automatic calculation of backpropagation. The kernel weights are initialized in the class as defined in Equation 6 using Kaiming [17] initialization, where the $S$ value and the kernel size are hyperparameters of the class.

        The Full FDC layer is an extension of the FDC layer that assumes the input images are in the spatial domain, so it will perform additional RFFT and Inverse RFFT (IRFFT) on the images. The Full FDC layer is created so that it can be used immediately in conventional CNN by swapping with any existing spatial convolution layers.

**Algorithm 2**: Full FDC layer

**Input:** Images with $C$ channels in spatial domain,
      $S$ number of $K_{height} \cdot K_{width}$ kernels in spatial domain
**Output:** Convoluted $C \cdot J$ channel images in spatial domain
**procedure** Full FDC layer
    Convert images into frequency domain with RFFT
    Input into FDC layer
    Convert images into spatial domain with IRFFT
    Remove convolution artifacts
    **return** convoluted $C \cdot J$ channel images
**end procedure**

### 2.4.2 *Frequency Domain Pooling (FDP)*
#### 2.4.2.1 *Frequency Domain Filtering*
Rippel et al. [7] utilized the technique of low pass filter to achieve frequency domain pooling by performing a square crop on the frequency image directly to remove higher frequencies and reduce image size. Although from the perspective of image processing, the technique may produce minor artifacts as it was a hard crop instead of a Gaussian mask. Their method is adapted to work with images transformed with RFFT as their method is simple and fast.

        Assuming the frequency domain image transformed with RFFT is already shifted in a way that the lower frequencies are concentrated in the middle left, while the higher frequencies are on the outer side (see Fig. 2b). In order to perform Frequency Domain Pooling (FDP), a rectangle hard crop is done on the frequency domain image by removing the same number of rows or columns from the top, bottom and right side of the image, so that the width and height ratio is maintained. Compared to spatial domain pooling, FDP has more flexibility in reducing the size, as the kernel size for max pooling is at least 2x2.

#### 2.4.2.2 *RFFT Shift*
In a RFFT image, the lower frequencies are located on the top left and bottom left. In order to shift the frequencies like FFT shift, the RFFT image is first split in half, then they are swapped (see Fig. 2). The technique is referred to as RFFT shift. The RFFT shift function is only used so that FDP can be done in a simpler way. It should be noted that RFFT shift is only needed to be performed once after RFFT and before IRFFT on the image. The shifted frequency domain image will not affect frequency domain convolution.

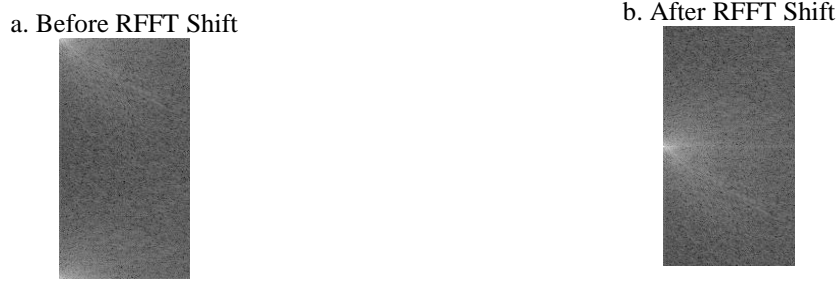

**Fig. 2** The magnitude spectrum representation of an image in the frequency domain with RFFT. The magnitude spectrum is used to visualize frequency domain images.

### 2.4.3 *Frequency Domain Activation Function*

As described previously, the absence of frequency domain activation function has been the biggest obstacle in the development of frequency domain CNN. The Second Harmonics SReLU (2SReLU) is an attempt in efficient frequency domain activation function by Watanabe and Wolf [13]. The 2SReLU activation function works by adding back the lower frequencies into the second harmonics, which is shown to be able to approximate the effect of ReLU in frequency domain. However, in our experiments, the results of using 2SReLU was not close to the effect of ReLU, instead, it creates double image artifacts as the lower frequencies are added back into the image. In the end, ReLU is still chosen to be used in frequency domain CNN with some constraint which will be discussed in Section 2.5.1.

### 2.5 *Frequency Domain CNN*
### 2.5.1 *Frequency Domain CNN (FDCNN) Architecture*

With the frequency domain components and implementation details described in Section 2.4, the proposed frequency domain CNN architecture which is referred to as FDCNN can now be defined. In order to take advantage of the frequency domain, the amount of domain transform for the images should be minimized. As the FDC and FDP layers are already defined to work directly with frequency domain images, the FDCNN would only require to perform domain transform twice on the image with RFFT and IRFFT, which is performed before and after all the FDC and FDP layers.

Due to the fact that there is still no suitable activation function for the frequency domain, and to take advantage of the fact that the images will be transformed back into the spatial domain for artifacts removal, a ReLU activation function can be added between the IRFFT and the fully connected layers (See Fig. 3).

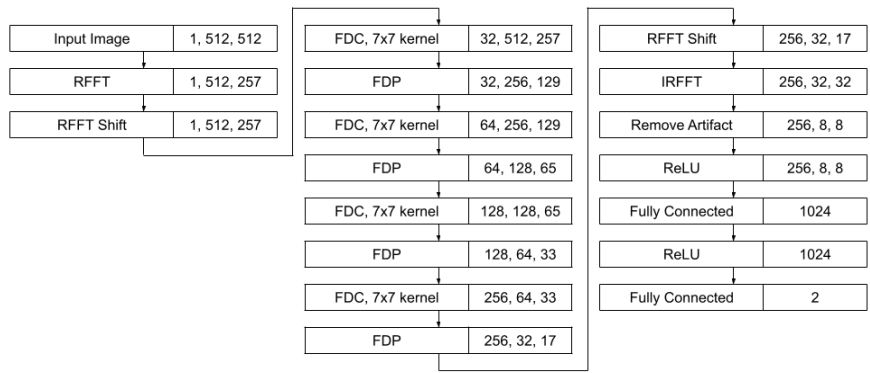

**Fig. 3** The proposed FDCNN architecture. The numbers are the output tensor shape arranged as channel numbers, height and width. The batch numbers are not included for simplicity.

The amount of FDC and FDP layers and its configurations can be adjusted to suit the specific need of the task. The experiment results from Section 3.2 and 3.3 can also be referred to when configuring the layers. For the APTOS 2019 diabetic retinopathy dataset, through various trial and error of adjusting the configurations and hyperparameters, the proposed FDCNN architecture illustrated in Fig. 3 appeared to work the best.

**2.5.2** *VGG16 with Full FDC Layers*
The VGG16 is a popular CNN architecture that is still commonly used, the architecture is shown to be highly capable for its achievements in the ImageNet Large Scale Visual Recognition Challenge (ILSVRC) in 2014. From Section 3.2, it is shown that the Full FDC layer is 46% faster than conventional convolution at 256/512 channels. For the VGG16 architecture as shown in Table 2, the conventional convolution layer that is configured with 256/512 channels is the first layer in Conv Block 4, the layer can be replaced with a Full FDC layer and observe speed improvements. Additionally, Conv Block 4 has another 2 conventional convolution layers configured with 512/512 channels, which can also be replaced with Full FDC layers.

**Table 2** The Conv Block 4 of VGG16 architecture and 2 modified VGG16 architecture with Full FDC layers. The numbers following Conv2d and Full FDC are the kernel size and input channels/output channels. The ReLU activation function is used after every Conv2d and Full FDC layer.

|  | **VGG16** | **VGG16-1FullFDC** | **VGG16-3FullFDC** |
|---|---|---|---|
| **Conv Block 4** | Conv2d, 3x3, 256/512<br>Conv2d, 3x3, 512/512<br>Conv2d, 3x3, 512/512<br>Max Pooling | Full FDC, 3x3, 256/512<br>Conv2d, 3x3, 512/512<br>Conv2d, 3x3, 512/512<br>Max Pooling | Full FDC, 3x3, 256/512<br>Full FDC, 3x3, 512/512<br>Full FDC, 3x3, 512/512<br>Max Pooling |

## 3 Results and Discussions
**3.1** *Experimental Setting*
The experiments are run on PyTorch version 1.8.0 with NVIDIA Tesla T4 GPU. The FDC and FDP layer always have the input image/tensor already in frequency domain, while Full FDC and spatial convolutions have the input image/tensor in the spatial domain.

The FDC layer is compared against PyTorch's spatial convolution layer Conv2d. It should be noted that when experiments are run on a NVIDIA GPU, the NVIDIA CUDA Deep Neural Network (CUDNN) library will be enabled by default. The CUDNN library is exclusively optimized on NVIDIA GPUs and it provides their own highly tuned implementations of the convolution layer. It must be emphasized that when CUDNN is enabled, every time when the Conv2d function is called, Conv2d is redirected to use one of NVIDIA's own spatial convolution depending on the Conv2d hyperparameters, NVIDIA's implementations include Winograd convolution and FFT-based convolution. Only when CUDNN is turned off, the Conv2d function will execute PyTorch's implementation of the classic spatial convolution. The CUDNN library is not an open-source project, and there is currently no way of specifying or knowing which convolution algorithm is used every time when Conv2d function is called.

For all CNNs, the experiments are run with CUDNN enabled which is the default setting, therefore conventional convolutions are used on conventional CNNs where Conv2d function is used. Note that FDCNN is not affected by CUDNN as the FDC layers are self-written and do not use the Conv2d function. The hyperparameters are as follows, the image size is 512x512, batch size is 4, the loss function is Cross Entropy Loss and the optimizer is Adam with learning rate of $1 \times 10^{-5}$ and weight decay of $1 \times 10^{-6}$.

In order to measure the performance of the FDCNN architecture defined in Section 2.5.1, an equivalent CNN architecture is created in a way that is structured similarly to common

CNN architectures while having similar configurations as the FDCNN (See Fig. 4). Similar to the FDCNN architecture, the CNN architecture has 4 convolution layers of the same configurations, the max pooling layers and fully connected layers also have the same configurations. The difference between the 2 CNN architectures is that CNN has ReLU activation function after every max pooling layer, which is a common structure but couldn't be done on FDCNN.

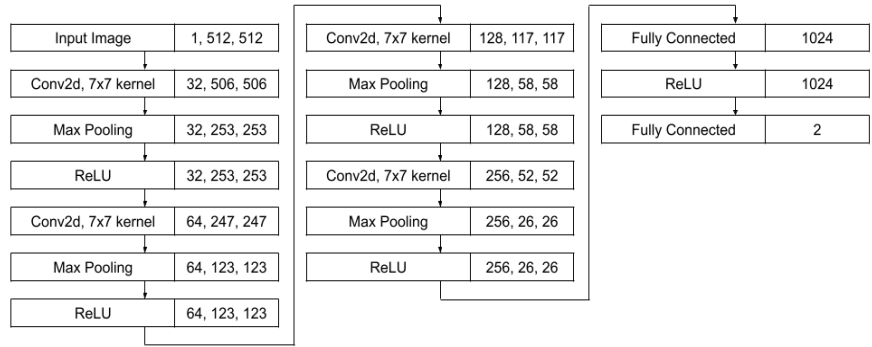

**Fig. 4** The CNN architecture to be compared with FDCNN architecture.

### 3.2 *FDC and Full FDC Layer*

The experiments are run to show the effect of the different hyperparameters on the performance of different convolution layers. The channel numbers, kernel sizes and image sizes are investigated on FDC, Full FDC, conventional convolution (Conv2d CUDNN ON) and spatial convolution (Conv2d CUDNN OFF) layers.

The first experiment is to investigate the effect of channel numbers on the execution time. From Fig. 5, it can be observed that FDC and conventional convolution have a similar performance on low channel numbers. Starting from 32/64 channels, FDC would be faster than conventional convolution, and the gap widens as the channel number increases. FDC is slower on lower channel numbers due to the overhead of the domain transformation on kernels. However, the benefits of CIC are more apparent as channel number increases because it reduces the number of convolutions to perform.

The Full FDC has a similar trajectory compared to FDC but overall slower. The reason is that Full FDC performs domain transformation twice on the image, which significantly increases the overhead. The Full FDC did perform better than spatial convolution starting from 64/128 channels, but it would take at least 256/512 channels to be faster than conventional convolution. It is positive that Full FDC can still be faster than conventional convolution, to be specific, Full FDC is 46% faster than conventional convolution at 256/512 channels, which would indicate that the Full FDC layer can replace conventional convolution layer that is configured with at least 256/512 channels in a conventional CNN and observe speed increase.

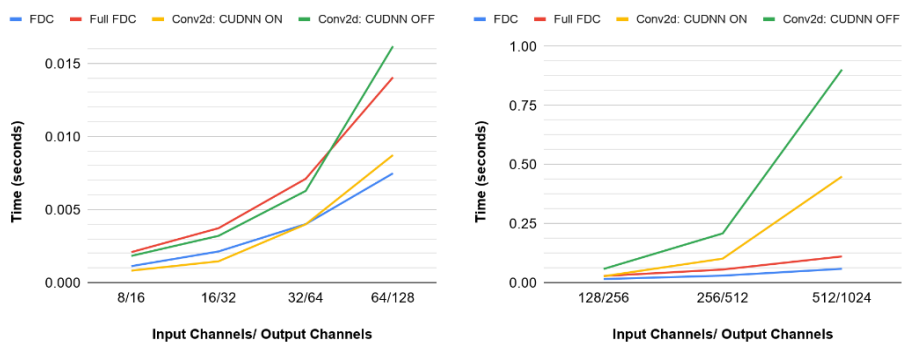

**Fig. 5** The effect of input channels and output channels on the execution time of different convolution layers. The input tensor is fixed at 512x512, and the kernel at 3x3.

The benefit of frequency domain convolution is that convolutions are pointwise products of the images and kernels. As the kernels are always the same size as the images after transforming into the frequency domain, the kernels that are defined in the spatial domain will have no effect on the execution time, which can be observed in Fig. 6. However, for spatial and conventional convolution, the execution time increases according to the kernel size. Starting from 5x5 kernel with 8/16 channels, both FDC and Full FDC are faster than spatial and conventional convolution.

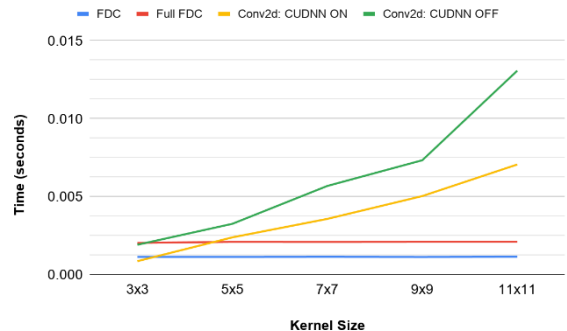

**Fig. 6** The effect of kernel size on the execution time of different convolution layers. The input tensor is fixed at 512x512, the input channels at 8, and output channels at 16.

The last experiment is to investigate the effect of image sizes. From Fig. 7, it is observed that the image size is not a factor that would affect the ranking of different convolution layers. Instead, the channel numbers and kernel size have a larger influence over the performance difference between convolution layers.

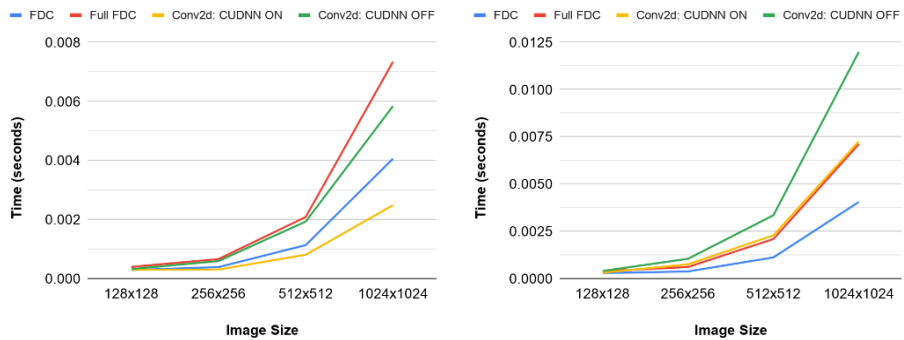

**Fig. 7** The effect of image size (input tensor) on the execution time of different convolution layers with different kernel size. The kernel size on the left figure is 3x3, while it is 5x5 on the right figure. Both the left and right figure has the input channels fixed at 8, and output channels at 16.

### 3.3 *FDP Layer*
The experiment is run to investigate the effect of image size on the speed of FDP, PyTorch's max pooling and average pooling layer. As discussed in Section 2.4.2.1, FDP can be performed very efficiently by removing rows and columns from the image, which leads to a constant time complexity (see Fig. 8). FDP is measured to take $3.17 \times 10^{-6}$ seconds.

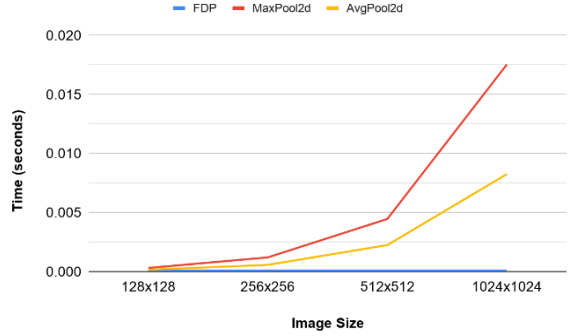

**Fig. 8** The effect of image size on the execution time of different pooling layers. All pooling layers are set to reduce the image size by half.

### 3.4 *FDCNN for DR Classification*

The experiment was first run for 40 epochs. Fig. 9 showed both CNNs are gradually overfitting as the number of epochs increased. The CNN is shown to reduce training loss much faster than FDCNN, possibly due to the larger image after convolutions, but its validation loss seemed relatively unstable. FDCNN is shown to plateau between 15 to 20 epochs before it overfits, so another experiment is run with training over 15 epochs.

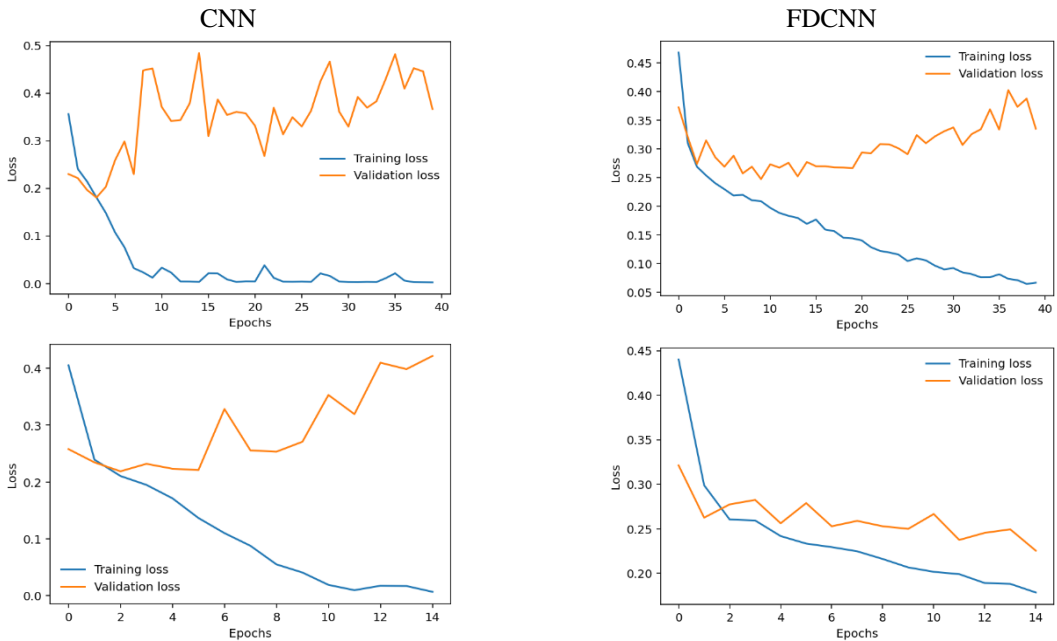

**Fig. 9** The loss graph of CNN and FDCNN over 40 and 15 epochs.

Although FDCNN has a lower validation loss at the end of 40 epochs, the CNN achieved 2.19% higher accuracy and a higher AUC (see Table 3). On the other hand, FDCNN is 54.21% faster and uses 70.74% less memory compared to CNN, which was contributed by the FDC and FDP layers in reducing the architecture weights and the computational complexity. In contrast, in the experiment where FDCNN is yet to overfit (15 epochs), FDCNN is observed to have a higher accuracy and AUC, while also maintaining to be 55.17% faster and 70.85% more memory efficient compared to CNN.

Although the FDCNN model trained over 15 epochs is 1.37% lower in accuracy compared to the CNN model trained over 40 epochs, they both have a very similar AUC score at 0.964 and 0.965. The similar AUC score infers that FDCNN is as capable as CNN. The results in Table 4 show that all p-values are smaller than 0.05 and the differences were statistically significant.

Table 3 Performance metrics of CNN and FDCNN model trained over 40 and 15 epochs.

| Epochs | 40 | | 15 | |
|---|---|---|---|---|
| | CNN | FDCNN | CNN | FDCNN |
| **Accuracy** | 92.90% | 90.71% | 90.16% | 91.53% |
| **Precision** | 91.64% | 91.06% | 89.66% | 91.42% |
| **Recall** | 94.61% | 90.57% | 91.11% | 91.91% |
| **AUC** | 0.964 | 0.956 | 0.958 | 0.965 |
| **Training time (seconds)** | 6082.77 | 2785.34 | 2319.75 | 1040.02 |
| **Memory usage (GB)** | 9.33 | 2.73 | 9.33 | 2.72 |

Table 4 Results of the Wilcoxon signed rank test to measure statistical significance with N=10.

| Methods | p-value < 0.05 |
|---|---|
| CNN (40e) & FDCNN (40e) | 0.00694 |
| CNN (15e) & FDCNN (15e) | 0.01242 |
| CNN (40e) & CNN (15e) | 0.00512 |
| FDCNN (40e) & FDCNN (15e) | 0.02202 |

From the experiments, it is observed that CNN can achieve higher accuracy when it is trained longer with some overfitting, while FDCNN would have its accuracy decreased when it starts to overfit. The main limitation of FDCNN would still be the lack of frequency domain activation function, it may restrict FDCNN architectures to be expanded with a large number of layers as ReLU is commonly accompanied together with each convolution layer. However, the experiments did seem to suggest that it may not be necessary to use ReLU that often as the FDCNN architecture only used ReLU once after the last FDC layer. Nonetheless, the experiments showed that FDCNN is as capable as CNN and is able to achieve a similar accuracy while significantly reducing the time and memory.

### 3.5 *VGG16 with Full FDC Layers for DR Classification*
The training is run over 15 epochs and the loss graph is shown in Fig. 10. All VGG16 architectures have a similar loss graph, but both VGG16 architectures with Full FDC layers seemed to be able to reduce the training loss slightly faster than the original VGG16. At the end of 15 epochs, VGG16-1FullFDC has a lower validation and training loss, which is reflected on the test accuracy. The VGG16-1FullFDC achieved an impressive accuracy of 95.63% which is 3.14% higher than VGG16 (see Table 5). It is believed that the CIC method in the Full FDC layer may be able to create different variations of features compared to conventional convolution, which increases the capability of the architecture to achieve better results. The results in Table 6 show that all p-values are smaller than 0.05 and the differences were statistically significant.

        VGG16                    VGG16-1FullFDC             VGG16-3FullFDC

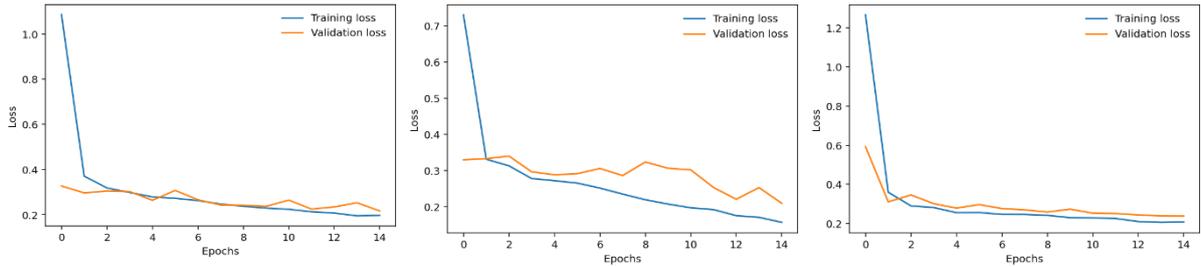

**Fig. 10** The loss graph of VGG16 and VGG16-1FullFDC over 15 epochs.

**Table 5** Performance metrics of VGG16 and VGG16-1FullFDC model trained over 15 epochs.

|  | **VGG16** | **VGG16-1FullFDC** | **VGG16-3FullFDC** |
| --- | --- | --- | --- |
| **Accuracy** | 92.49% | 95.63% | 94.40% |
| **Precision** | 94.89% | 94.49% | 96.09% |
| **Recall** | 90.03% | 97.04% | 92.72% |
| **AUC** | 0.917 | 0.940 | 0.950 |
| **Training time (seconds)** | 3803.92 | 3666.94 | 3517.00 |
| **Memory usage (GB)** | 8.65 | 8.76 | 8.89 |

**Table 6** Results of the Wilcoxon signed rank test to measure statistical significance with N=10.

| **Methods** | **p-value < 0.05** |
| --- | --- |
| VGG16 & VGG16-1FullFDC | 0.00512 |
| VGG16 & VGG16-3FullFDC | 0.00512 |
| VGG16-1FullFDC & VGG16-3FullFDC | 0.00694 |

The VGG16-1FullFDC architecture did achieve 3.60% faster in training time but also 1.27% higher in memory usage. It should be clear that the Full FDC layer is not as efficient as the FDC layer due to the immediate use of RFFT and IRFFT in the layer itself for it to work in the spatial domain. The increase in memory usage is believed to be contributed by the immediate domain transforms, where the GPU may be keeping the images of both domains in the memory, unlike FDCNN where the image in the spatial domain is not needed anymore after RFFT. Nonetheless, the VGG16-1FullFDC resulted in a lower training time which was predicted from the results of 256/512 channels Full FDC layer in Section 3.2.

The VGG16-3FullFDC also achieved 4.09% increase in speed and 1.48% increase in memory usage. Although the VGG16-3FullFDC showed 1.23% decrease in accuracy, it achieved a higher AUC score compared to VGG16-1FullFDC, which suggests that VGG16-3FullFDC may have a higher capability with the additional Full FDC layers. However, the additional Full FDC layers may have increased the complexity of the architecture, leading to a slower reduction in training loss, thus a higher number of epochs may be needed to further reduce the loss and achieve a higher accuracy.

To compare the results with CNN and FDCNN, the VGG16 architecture does have a much stable loss graph, probably contributed by the much robust configuration of the convolution layers. The larger number of layers also resulted in a longer training time in 15 epochs, specifically, VGG16 is 63.98% slower than CNN and 265.76% slower than FDCNN, but with only 0.96% increase in accuracy compared to FDCNN. On the other hand, VGG16-1FullFDC is better justified for its much higher accuracy.

## 4 Conclusion

The research proposed FDC and Full FDC layers constructed with RFFT, kernel initialization strategy, convolution artifact removal and CIC method. Together with the improved FDP layer, the layers are used to construct the frequency domain CNN architecture and to modify the existing VGG16 architecture. The FDC layer is shown to be faster than conventional convolution layer starting from only 32/64 channels, while it is at least 256/512 channels for the Full FDC layer. The FDP layer is also shown to be faster than max pooling and average pooling regardless of the image size. For diabetic retinopathy classification, the VGG16-1FullFDC architecture achieved an impressive accuracy of 95.63% while having a shorter training time than VGG16. The FDCNN architecture is also shown to be at least 54.21% faster and 70.74% more memory efficient compared to an equivalent CNN.

To conclude, the frequency domain components including FDC, Full FDC and FDP layers are shown to be faster than their spatial domain counterparts. The frequency domain components are also used to build frequency domain CNNs including FDCNN and VGG16-1FullFDC which are shown to be faster and are able to achieve higher accuracy compared to equivalent CNN architectures. A fast and efficient CNN architecture like FDCNN would not only be beneficial for image classification tasks, FDCNN could be used directly to accelerate other machine learning frameworks that relies on CNNs, for example the R-CNN for object detection and the Generative Adversarial Network (GAN) for image supersampling.

The research hopes to increase awareness and to accelerate the research and development of frequency domain CNN especially in discovering a suitable activation function. Even without frequency domain activation functions, FDCNN and VGG16-1FullFDC have demonstrated to have the potential for further improvements. A deeper or a more complex structure can be explored for FDCNN. The Full FDC layer can also be implemented in other popular CNN architectures.

**Disclosures**
No conflict of interest, financial or otherwise, are declared by the authors.

[10] Pratt, H. (2019). *Deep Learning for Diabetic Retinopathy Diagnosis & Analysis.* PhD thesis, University of Liverpool. doi:10.17638/03046567

[11] Vasilyev, A. (2015). CNN optimizations for embedded systems and FFT. *Stanford University Report*. Retrieved from http://cs231n.stanford.edu/reports/2015/pdfs/tema8_final.pdf

[12] Ayat, S. O., Khalil-Hani, M., Ab Rahman, A. A., and Abdellatef, H. (2019). Spectral-based Convolutional Neural Network without multiple spatial-frequency domain switchings. *Neurocomputing, 364*, 152-167. doi:10.1016/j.neucom.2019.06.094

[13] Watanabe, T., & Wolf, D. F. (2020). Image classification in frequency domain with 2SReLU: a second harmonics superposition activation function. *arXiv preprint arXiv:2006.10853.*

[14] Wang, Z., Lan, Q., Huang, D., & Wen, M. (2016). Combining FFT and spectral-pooling for efficient convolution neural network model. In *2016 2nd International Conference on Artificial Intelligence and Industrial Engineering (AIIE 2016)* (pp. 203-206). https://doi.org/10.2991/aiie-16.2016.47

[15] Gonzalez, R. C., & Woods, R. E. (2007). *Digital Image Processing (3rd Edition)*. Pearson.

[16] Oppenheim, A. V., Schafer, R. W., & Buck, J. R. (1999). *Discrete-Time Signal Processing* (2nd ed.). Upper Saddle River, N.J: Prentice Hall.

[17] He, K., Zhang, X., Ren, S., & Sun, J. (2015). Delving deep into rectifiers: Surpassing human-level performance on imagenet classification. In *2015 IEEE International Conference on Computer Vision (ICCV).* doi:10.1109/iccv.2015.123
16